\documentclass[11pt,a4paper]{article}

\usepackage{geometry}
\usepackage{algorithm}
\usepackage{algpseudocode}
\usepackage{tikz}
\usepackage{subcaption}
\usepackage{amsfonts}
\usepackage{mwe}
\usepackage{mathtools}
\usepackage{booktabs}
\usepackage[svgnames,table]{xcolor}
\usepackage[tableposition=above]{caption}
\usepackage{pifont}
\usepackage{colortbl}
\usepackage{tabularx}
\usepackage{adjustbox}
\usepackage{longtable}
\usepackage{float}
\usepackage{hyperref}
\usepackage{verbatim}
\usepackage{authblk}
\usepackage[T1]{fontenc}
\usepackage{textcomp}

\DeclarePairedDelimiter\ceil{\lceil}{\rceil}
\DeclarePairedDelimiter\floor{\lfloor}{\rfloor}

\usepackage{natbib}
 \bibpunct[, ]{(}{)}{,}{a}{}{,}%

\title{Speeding Up Mixed-Integer Programming Solvers with Sparse Learning for Branching}

\author[$\ast$]{Selin Bayramo\u{g}lu}
\author[$\ast$]{George L Nemhauser}
\author[$\ast,\dagger$]{Nikolaos V Sahinidis}
\affil[$\ast$]{H. Milton Stewart School of Industrial and Systems Engineering, Georgia Institute of Technology}
\affil[$\dagger$]{School of Chemical \& Biomolecular Engineering, Georgia Institute of Technology\par
\texttt{\{sbayramoglu3, gn3, nikos\}@gatech.edu}}

\date{}

\begin{document}
\maketitle

\begin{abstract}
Machine learning is increasingly used to improve decisions within branch-and-bound algorithms for mixed-integer programming. Many existing approaches rely on deep learning, which often requires very large training datasets and substantial computational resources for both training and deployment, typically with GPU parallelization. In this work, we take a different path by developing interpretable models that are simple but effective. We focus on approximating strong branching (SB) scores, a highly effective yet computationally expensive branching rule. Using sparse learning methods, we build models with fewer than 4\% of the parameters of a state-of-the-art graph neural network (GNN) while achieving competitive accuracy. Relative to SCIP's built-in branching rules and the GNN-based model, our CPU-only models are faster than the default solver and the GPU-accelerated GNN. The models are simple to train and deploy, and they remain effective with small training sets, which makes them practical in low-resource settings. Extensive experiments across diverse problem classes demonstrate the efficiency of this approach.
\end{abstract}

\section{Introduction}\label{sec1}

Mixed-integer programs (MIPs) arise in many real-world applications, including airline crew scheduling \citep{tdh22}, energy systems planning \citep{kow20}, drone-assisted parcel delivery \citep{pgw19}, and electric vehicle routing \citep{fjml22}. State-of-the-art MIP solvers typically rely on branch-and-bound \citep{ld1960}, supplemented with techniques such as cutting planes and primal heuristics. Over the past two decades, improvements to these components have led to major gains in performance: between 2001 and 2020, MIP solvers improved by roughly three orders of magnitude \citep{koch+22}. Even so, performance has remained highly sensitive to solver design choices, especially branching rules, which shape the search tree and strongly influence runtime. Although modern solvers such as SCIP have more than 1600 parameters, their default settings are tuned for general-purpose use and can be far from optimal in specialized applications.

The broad success of machine learning (ML) in science and engineering has motivated its use in optimization to speed up the solution of MIPs. Early work emphasized automated algorithm configuration \citep{hutter+10, xu+11}, aiming to capture interactions among configuration parameters. More recent efforts have focused on improving decisions made within the branch-and-bound process itself. Branching has received particular attention because it strongly affects the size of the search tree; poor branching choices can lead to exponential slowdowns. For example, \citet{a+13} showed that na\"ive branching strategies can increase solve times by more than a factor of eight, rendering otherwise tractable instances difficult to solve.

Much of this line of work has centered on strong branching (SB), a highly effective but computationally expensive branching algorithm that often produces small search trees \citep{achterberg07thesis}. To retain SB's benefits while reducing its cost, ML models have been developed to predict SB-related quantities, including candidate rankings \citep{khalil+16}, branching scores \citep{alvarez+17}, and the best branching candidate \citep{gasse+19}. Beginning with the work of \citet{gasse+19}, graph neural networks (GNN) have become the dominant approach. These high-capacity models, trained on large datasets, have consistently improved over standard solver behavior in settings where many similar MIPs are solved.

\subsection{Contributions}

In this work, we make three main contributions. First, we show that simple, interpretable models for predicting SB scores can speed up a state-of-the-art MIP solver across four widely studied classes of combinatorial optimization problems. Although our models use no more than 4\% of the parameters of a leading GNN model, they consistently reduce solve times compared to both default SCIP and GNN-based branching. Because they build on established feature sets from the literature and remain simple by design, they are easier to implement and more practical to deploy.

Second, whereas many GNN-based approaches rely on GPU acceleration to be competitive, our models are trained and deployed entirely on CPUs, which allows seamless integration into modern solvers without specialized hardware. In a direct comparison, we find that our CPU-only models perform better even when the GNN baselines run on GPUs. Unlike hybrid GNN methods \citep{gupta+20}, our approach does not depend on deep learning components.

Third, we demonstrate that our models remain effective even with substantially smaller training sets. In contrast to prior deep learning approaches (\citet{gasse+19}, \citet{gupta+20}, \citet{gupta+22}) that depend on large-scale data collection, our models achieve comparable performance with far fewer samples. This data efficiency makes them practical for settings where generating training data is expensive or impractical, and it broadens the reach of ML-driven branching in real applications.

\section{Background Material}

We consider MIPs of the following form:
\begin{equation}
\left.
\begin{aligned} 
\min_{x} \quad
& c^T x\\
\text{s. t.}\quad
& Ax \leq b\\
& l \leq x \leq u\\
& x_j \in \mathbb{Z} \quad j \in I = \{1,\ldots,p\} \\
& x_j \in \mathbb{R} \quad j \in \{p+1,\ldots,n\}, \\
\end{aligned} \quad \quad \right\}
\label{mip}
\end{equation}
where $c \in \mathbb{R}^n$ is the vector of objective coefficients, $A \in \mathbb{R}^{m \times n}$ is the constraint matrix, $b \in \mathbb{R}^m$ is the right-hand-side vector, $l, u \in \mathbb{R}^n$ are the lower and upper bounds for variables, and $x \in \mathbb{Z}^p \times \mathbb{R}^{n - p}$ is the vector of decision variables, $p$ of which are restricted to take integral values.  We denote the set of integer variables in $x$ as $x_I$.

Branch-and-bound is the core algorithm behind modern MIP solvers. It starts by solving a relaxation of the original problem, typically the linear programming (LP) relaxation obtained by removing integrality constraints. The optimal relaxed solution provides a valid lower bound on the optimal objective value. If the solution $\hat{x}$ is integral ($\hat{x}_I \in \mathbb{Z}^p$), the algorithm terminates. Otherwise, the solver branches by selecting a fractional variable $x_j$, and creating two subproblems with added bounds $x_j \leq \floor*{\hat{x}_j}$ and $x_j \geq \ceil*{\hat{x}_j}$. Each subproblem is then solved in the same way, and the procedure is then repeated recursively. For a detailed treatment, see \citet{nw99book}.

\subsection{Branching}

Branching is central to branch-and-bound because it helps shape the search tree and largely determines its size, which directly affects solve time. Modern MIP solvers include many branching strategies, and \textit{strong branching} is among the most effective and widely studied.

\textbf{Strong Branching}. Strong branching (SB) was introduced by \citet{applegate+98} in the context of the traveling salesman problem. SB scores candidate variables by explicitly simulating the branching step. Consider a subproblem $N$ with LP relaxation value $z$ and optimal solution $\hat{x}$, and let ${x}_j$, $j \in I$, be fractional in $\hat{x}$. Define $N_j^-$ and $N_j^+$ as the LPs obtained by adding the constraints $x_j \leq \floor*{\hat{x}_j}$ and $x_j \geq \ceil*{\hat{x}_j}$, respectively. Assuming both are feasible, let $z_j^-$ $ (z_j^+)$ be the optimal LP value of the down (up) child LP. The SB score is a function of $z_j^- - z$ and $z_j^+ - z$, the objective gains in both directions. The state-of-the-art scoring function is the product function,
\begin{equation}
s_j^P = \max\{z_j^- - z, \epsilon\} \times \max\{z_j^+ - z, \epsilon\}\label{eq:score}
\end{equation}
where $\epsilon$ is a small number (for example, $10^{-6}$) \citep{achterberg07thesis}. If a child is infeasible, the node is pruned and $z_j^-$ (or $z_j^+$) is set to a very large value. The candidate with the highest score is then selected for branching. 

Empirically, SB often yields small search trees \citep{achterberg07thesis}, but it is expensive: scoring a single candidate requires solving two child LP relaxations at each node. In practice, solvers reduce this cost by applying SB only to a short list of promising candidates or by limiting the number of simplex iterations during the child LP solves.

\textbf{Pseudocost branching}. Pseudocosts, introduced by \citet{benichou+71}, offer a cheaper alternative to SB by leveraging information from previous branching decisions. The down (up) pseudocost of a variable measures the average improvement in the dual bound per unit change in the variable's LP value when branching down (up). Let $\phi^-$ and $\phi^+$ be the down and up pseudocosts of an integer variable $x_j$ and suppose $\hat{x}_j \notin \mathbb{Z}$. Then, the change in the dual bound for the down child is estimated by $(\ceil{\hat{x}_j} - \hat{x}_j) \phi^-$. For the up child, this estimate is $(\hat{x}_j - \floor{\hat{x}_j}) \phi^+$. The pseudocost score is then calculated by Equation~\eqref{eq:score}, replacing true bound changes with these estimates. Since pseudocosts approximate SB outcomes, pseudocost branching (PB) is typically much cheaper than SB, though also less accurate.

To combine the accuracy of SB with the efficiency of PB, solvers often use hybrid strategies. \textit{Reliability branching} uses pseudocosts for ``reliable'' variables, those branched on often enough to yield meaningful estimates, while applying SB to unreliable variables until a stopping criterion is satisfied. \textit{Hybrid branching}, the default strategy in many modern MIP solvers, builds on reliability branching by incorporating additional measures, such as the number of subproblems pruned or the number of bound reductions triggered by a candidate variable \citep{ab09}.

Beyond these classical methods, researchers have explored ways to incorporate richer information into branching decisions. For example, \citet{bs13} proposed \textit{cloud branching}, which evaluates candidates across multiple optimal LP solutions rather than a single solution, and \citet{gl11} studied deeper branching rules that evaluate variables using dual-bound improvements not only at the child nodes but also two levels down the tree.

Researchers have also explored alternative branching criteria. For example,  \citet{pc07} selected variables based on their impact on active LP constraints to help guide the search toward feasibility, while \citet{k+09} used information from fathomed nodes in incomplete searches to prioritize variables that are more likely to produce additional pruning. Theory has complemented these empirical efforts: \citet{ln17} studied branching in an abstract MIP setting and derived rules that outperformed a state-of-the-art solver in both node count and runtime.

Several recent studies also revisit standard assumptions about what makes a good branching candidate. While solvers typically branch only on fractional integer variables, \citet{dey+24} demonstrated that branching on integer-valued variables, in some cases, reduces the search tree. More broadly, branching need not be limited to single-variable disjunctions. Work on general disjunctions, i.e., branching on $\pi^T x \leq \pi_0$ and $\pi^T x \geq \pi_0 + 1$ with integer coefficients, has shown promise; for instance, \citet{yang+21} found that disjunctions on sums of variables can substantially shrink search trees for 0-1 knapsack problems. Together, these results expand the range of branching strategies and point to additional opportunities for improving solver performance.

\section{Machine Learning for Branching}\label{litrev}

Machine learning is now widely used across scientific fields, including integer programming. Recent work has applied ML to accelerate MIP solving by improving decisions within branch-and-bound solvers, such as node selection \citep{h+14, yys21}, primal heuristics \citep{khalil+17, nair+21, sonnerat+21, chmiela+21}, and cutting plane selection \citep{tang+20, paulus+22}. Among these components, \textit{branching} has attracted the most attention. A common thread is supervised learning to approximate strong branching (SB), which is highly effective but expensive and relies only on local information at the current node. Since SB can substantially reduce tree size yet becomes costly to apply broadly, learning fast approximations has emerged as a practical route to improving solver performance.

An early study in this direction is \citet{khalil+16}, who trained a ranking SVM to separate \textit{good} from \textit{bad} branching candidates using SB scores as labels. Their approach is an \textit{instance-specific online approach}: during the early phase of solving a given instance, they collect data to train a ranking model tailored to that instance, and then use it to guide subsequent branching decisions. On the MIPLIB 2010 benchmark \citep{koch+11}, their method was competitive with a commercial state-of-the-art solver.

Building on this idea, \citet{alvarez+17} predicted SB scores directly using Extremely Randomized Trees (ExtraTrees) and reported strong results on both synthetic and benchmark instances. In related work, \citet{alvarez+16tr} proposed an online regression model that predicts SB scores for reliable variables while applying SB to unreliable ones, mirroring reliability branching; the model updates as more data became available, improving over the course of the solve. Similarly, \citet{gu+22} used ExtraTrees to predict SB scores for security-constrained unit commitment and found that finer variable groupings improved performance.

More recently, \citet{yang+22} introduced \textit{generalized strong branching}, which extends SB from single variables to sets of variables. They trained gradient-boosted decision tree models to rank candidate sets and showed that this approach outperformed a state-of-the-art solver.

Finally, \citet{ghaddar+23} extended learning to branch to nonlinear problems. They trained several models to select the best branching rule from a portfolio of branching rules for spatial branching, that is, branching on continuous variables. The learned rules outperformed standard ones on benchmark instances from the literature.

\subsection{Imitation learning}
Another influential research direction is imitation learning, which trains a branching policy to mimic an expert. Typically, a large set of instances from a problem family is generated, and the model is trained to reproduce SB decisions on this data. The resulting policy is tailored to that family and aims to match expert branching behavior at much lower computational cost.

A seminal example is by \citet{gasse+19}, who trained graph convolutional neural networks (GNNs) to imitate SB on four NP-hard problems: set covering, combinatorial auctions, maximum independent set, and capacitated facility location. They demonstrated that a GNN-based branching rule, when evaluated on a GPU, outperformed the default SCIP solver and several alternatives, and it generalized to larger problem sizes than those used for training.

\citet{gupta+20} later noted that the performance of GNN models trained in \citet{gasse+19} was highly dependent on GPU acceleration; when the same GNN models were run on CPUs, the default solver was faster. To reduce inference cost, they proposed \textit{hybrid architectures} that apply a GNN only at the root node and switch to cheaper models (MLPs or SVMs) deeper in the tree.  While this trades off some accuracy, it yielded faster solve times than both SCIP and pure GNNs in CPU-only environments. In follow-up work, \citet{gupta+22} observed that the best SB candidate at a node often matches the parent's second-best candidate, and they encoded this bias into training by modifying the loss function to favor such candidates, which improved generalization of the learned policy.

Other studies have expanded the scope of imitation learning for branching. \citet{nair+21} trained deep models to imitate a scalable SB variant and reported improvements over SCIP in both primal-dual gap and runtime on large instances and MIPLIB benchmarks.  \citet{zarpellon21} tackled heterogeneous MIPs, where instances span multiple families, by combining features of variables and search-tree state and using hybrid branching as the expert, producing models that generalized across varied classes.

\begin{figure}[ht!]
    \caption{A common ML pipeline used in previous studies (top path) and our approach (bottom path) for learning to branch.}
    \includegraphics[width=\textwidth]{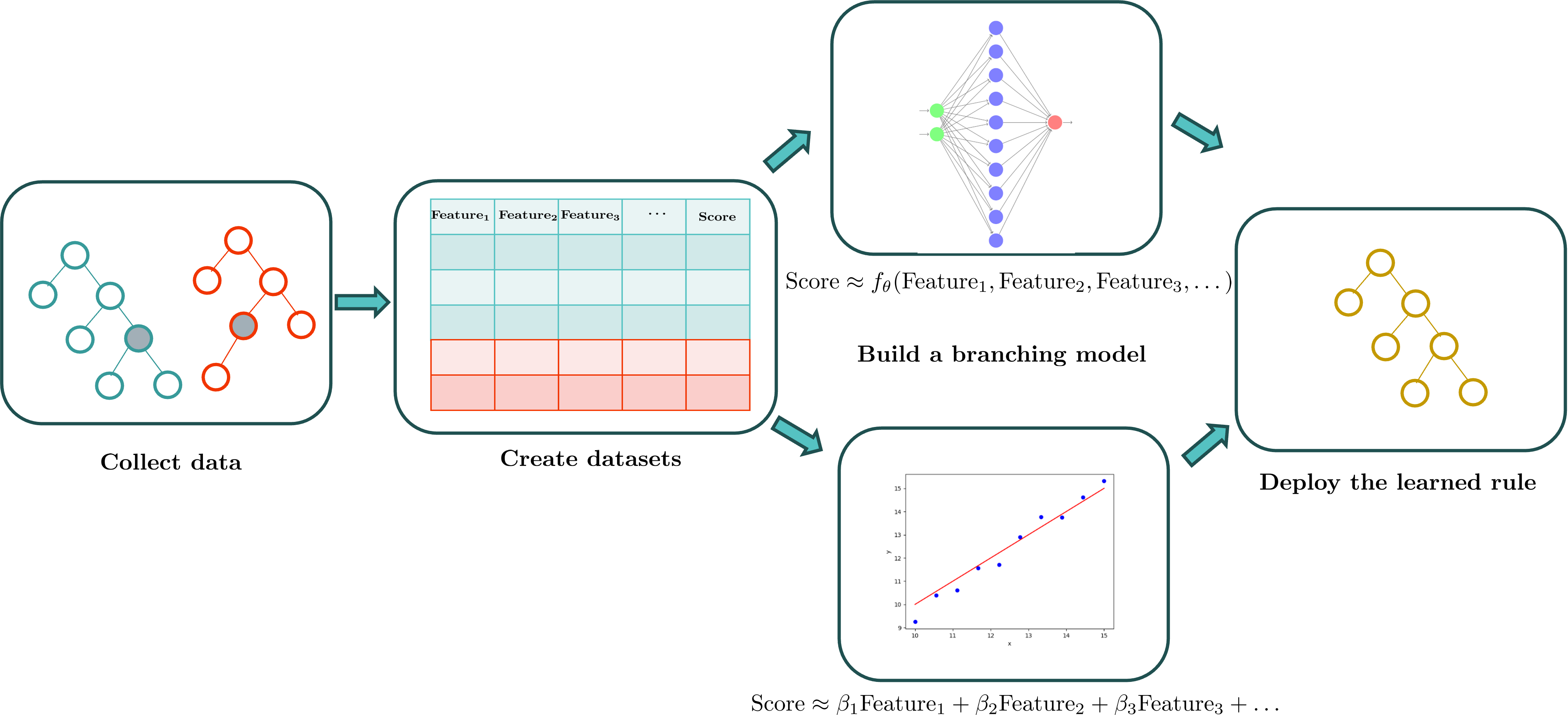}
    \label{fig:workflow}
\end{figure}

Figure~\ref{fig:workflow} summarizes the common workflow in imitation-learning studies and contrasts it with our approach. The typical pipeline is as follows: (1) solve many instances while computing SB (or another expert) scores; (2)  collect features and labels, often at high computational cost; (3) train a model to approximate the expert scores; and (4) deploy the model to replace expensive score computation during solving. Neural network methods follow the top path, producing large black-box models that typically benefit from GPU acceleration. Our approach follows the same overall pipeline but trains sparse linear models on quadratic feature transformations, yielding compact predictors with few parameters.

\subsection{Reinforcement learning}
A common drawback of imitation learning is that it is tied to a fixed expert policy. When the expert is suboptimal for a given problem class, the learned policy inherits these weaknesses. To reduce this dependence, researchers have explored reinforcement learning (RL), which formulates branching as a Markov Decision Process (MDP). In this setting, the solver learns branching behavior through trial and error and can, in principle, discover strategies beyond traditional expert policies.
\citet{etheve20} and \citet{s+22} applied deep RL to branching with the goal of minimizing the size of the subtree rooted at any node. They also identified conditions under which this local objective is consistent with minimizing the overall search tree, a useful proxy for B\&B efficiency. In related work, \citet{sun+20} used an evolutionary strategy to train an RL agent and reported consistent improvements over competing methods across several problem families, often achieving the best solve times.

\section{Methodology}

\subsection{Problems}

We evaluate our approach on four NP-hard problem domains that are widely used for studying learning branching rules: set covering (SC), combinatorial auctions (CA), maximum independent set (IS), and capacitated facility location (FL). These domains were introduced in this context by \citet{gasse+19} and have since been used in follow-up work (\citet{gupta+20, gupta+22}, and \citet{s+22}). 

For each domain, we generate 10,000 training instances, along with 2,000 instances each for validation and testing. We use the validation sets to tune hyperparameters for our models and all baselines, and we use the test sets to measure how accurately each method approximates SB decisions. To evaluate solver performance, we follow \citet{gasse+19} and create 20 small, 20 medium, and 20 large problem instances per domain, which we solve with SCIP under different branching strategies. Problem sizes and generators match those in \citet{gasse+19} to enable direct comparison.

\subsection{Features}

Our models use sparse feature representations that build on the feature set of \citet{khalil+16} and the aggregated variable features of \citet{gasse+19}, with several key modifications.

Our work considers as dynamic many features previously considered static. In \citet{khalil+16}, features computed at the root node (static features) are reused at all subsequent branching nodes. Static features include data from the root LP, such as how often a variable appears across constraints and summary statistics of constraint coefficients. Instead, we recalculate these quantities at every branching step so they track the current LP relaxation. As constraints (original rows or cutting planes) are added or removed, and variables are fixed during the search, we update the associated counts and statistics at each node of the search tree. 

While \citet{gasse+19} represented MIPs as bipartite graphs with features for variables, constraints, and edges, our models use variable features only. To retain relational information without the full graph, we summarize each candidate's neighboring constraint and edge features using the mean, maximum, and minimum. 

Following \citet{sun+20}, we remove fixed variables at each node to form a reduced formulation and calculate features on this simplified representation. This adjustment provides a cleaner and more accurate description of the active problem structure.

Table~\ref{tbl:features} in the Appendix lists all features used in this study.

\subsection{Building sparse models}

We study whether sparse linear or quadratic models over branching features can adequately capture SB. To encourage sparsity without sacrificing predictive accuracy, we use lasso regularization \citep{tibshirani96}. Specifically, we solve:
\begin{align} 
\min_{(\beta, \beta_0)} \quad
& \frac{1}{2} || y - \beta_0 - X \beta||^2_2 + \lambda || \beta ||_1 \label{sparse_opt}
\end{align}
Assume the training set contains $k$ branching candidates and $p$ features. Let $y \in \mathbb{R}^k$ denote the vector of normalized SB scores and $X \in \mathbb{R}^{k \times p}$ the feature matrix. The decision variables are the feature coefficients $\beta \in \mathbb{R}^p$ and the intercept $\beta_0 \in \mathbb{R}$. The hyperparameter $\lambda \geq 0$ penalizes the $l_1$-norm of the coefficient vector $\beta$. Solving problem (\ref{sparse_opt}) for a value of $\lambda$ determines a model ($\hat{\beta}, \hat{\beta_0}) \in \mathbb{R}^{p+1}$, that is, $\hat{y} = \hat{\beta_0} + X\hat{\beta}$.

Lasso improves interpretability while maintaining predictive performance through $l_1$ regularization; penalizing the $l_1$-norm of the coefficient vector drives many coefficients to zero, leaving a smaller set of informative features.

We train the lasso models using the R-package \texttt{glmnet} \citep{friedman+10}. The package computes the lasso regularization path (solving the problem over a sequence of automatically chosen values of $\lambda$ and returning the corresponding coefficient vectors $\beta$). We then select the model that yields the lowest mean-squared error (MSE) on the validation set.

\subsection{Data collection}\label{data}

All experiments use SCIP 8.0.0 \citep{scipoptsuite8} with SoPlex 6.0.0 as the LP solver. To maintain consistency across runs, we disabled presolve restarts and cutting-plane generation at non-root nodes.

Following \citet{gasse+19}, at each subproblem we either apply strong branching to all candidates or not at all, selecting the former option with probability 5\%, in which case we collect data. Let $C$ denote the set of branching candidates at a node. When SB is executed, we store an SB data tuple containing: ($i$) the GNN input features in terms of a bipartite graph representation of the node state, $\textbf{G} = (G, \text{C}, \text{V}, \text{E})$, where $G$ is the variable-constraint graph, $\text{C} \in \mathbb{R}^{m \times d_C}$ is the matrix of constraint features, $\text{V} \in \mathbb{R}^{n \times d_V}$ is the matrix of variable features, and $\text{E} \in \mathbb{R}^{m \times n \times d_E}$ is the matrix of edge features; ($ii$) $\mathbf{i}^*_{\text{SB}} \in C$, a candidate with the highest SB score, ($iii$) branching candidate features $\textbf{X} \in \mathbb{R}^{|C| \times d}$, and ($iv$) normalized SB scores $\textbf{s} \in \mathbb{R}^{|C|}$, obtained by dividing by the $l_2$-norm of the score vector. 
We refer to $(\textbf{G}, \mathbf{i}^*_{\text{SB}}, \textbf{X}, \textbf{s})$ as the SB data tuple. 

Unlike \citet{gasse+19}, who include all free integer variables as candidates, we restrict candidates to fractional integer variables. In practice, variables that are already integral in the LP solution tend to receive small SB scores and can be omitted with little loss. For instance, if a binary variable $x_j$ takes value 0 in the LP solution, branching down fixes it to 0 again, leaving the objective unchanged ($z_j^+ = z$). The resulting product score in (\ref{eq:score}) is therefore close to zero, making $s_j^P$ uninformative.

For data collection, we use SCIP's \textit{vanilla full strong branching} (VFS) with the non-default settings \texttt{scoreall = TRUE} and \texttt{idempotent = TRUE}. Relative to SCIP's default behavior, these options (a) ignore primal solutions found during SB and (b) force SB to evaluate all candidates, even when branching yields an infeasible subproblem. Finally, to ensure data quality, we discard any node where SB triggers numerical issues and SCIP reports errors.

\subsubsection{Large-sample setting}

We consider two approaches for generating training data. The first, which we call the \textit{large sample setting}, mirrors \citet{gasse+19}: we build a large dataset by solving many instances and recording SB data. To increase diversity, we cap the number of processed nodes per instance at 10,000 for FL and 100,000 for the other domains. Using this procedure, we generate 100,000 SB data tuples for training, and 20,000 tuples each for validation and testing.  In all cases, instances are sampled uniformly at random with replacement, and each selected instance is solved under multiple SCIP random seeds.

We use the full tuple collection above to train GNNs. For consistency with prior work \citep{gasse+19}, we then downsample the tuple pool to form candidate-level datasets: 250,000 candidate observations for training and 100,000 for each of validation and testing. This reduction can be substantial in domains with many candidates; for example, in the independent set domain, we generate roughly 43 million training candidates and retain about 0.6\%.

Finally, for each problem domain, we create five independent training datasets by varying the random seed used to select SB data tuples. This procedure yields multiple distinct training sets, reducing the chance that results depend on a single sample.

\subsubsection{Small-sample setting}

The second data-collection scheme explores the simplicity of sparse models and aims to reduce the number of SB measurements. Although Table~\ref{tbl:features} lists 101 features, preprocessing typically removes around 30\% of them, leaving a much smaller active set. Even with quadratic interactions included, the resulting models involve only a few thousand terms. Based on this observation, we estimate 25,000 candidate-level measurements as sufficient for training. 

Since this scheme uses orders of magnitude fewer measurements than the large sampling setting, we refer to it as the \textit{small-data setting}. Its lower cost also makes it feasible to build datasets separated by problem size, which is often impractical for deep learning models that require substantially larger training sets. 

For each problem domain and size, we generate separate training and validation datasets by sampling 300 instances per size and solving them until we collect 25,000 candidate observations for each split. To increase measurement diversity, we enforce tighter caps than in the large sample setting; we limit processing to 1000 nodes per instance so that no single problem dominates the dataset. In addition, for medium and large IS instances (where a single node can yield more than 800 candidate observations), we cap the number of collected measurements per instance at 5000. For most domains, SB is applied with 5\% probability, consistent with the large-sample scheme; for FL problems, we apply SB at every node because otherwise each instance yields too few samples.

As in the large-sample setting, we generate five independent training datasets for each problem domain and size, along with a separate validation set. Each dataset is produced by solving the same pool of instances in a different randomized order.

Table~\ref{tbl:dataset-breakdown} summarizes the data-generation cost under the large-sample (LS) and small-sample (SS) settings. For each configuration, we report (i) the average number of instances solved and (ii) the average number of subproblems generated. Row 1 corresponds to GNN sampling. Row 2 reports the large-sampling scheme for sparse models. Rows 3-5 report the small-sampling scheme for sparse models at small, medium, and large problem sizes, respectively.

\begin{table}[ht]
\caption{Data collection statistics for training datasets with different sampling schemes. The first entry is the average number of instances solved and the second entry is the average number of subproblems solved to generate the dataset.}
    \centering
    \begin{tabular}{l c c c c}
        \toprule
        Data & SC & CA & FL & IS \\
        \midrule
        GNN data & 7617 (100K) & 11472 (100K)  & 6193 (100K) & 5429 (100K) \\
        LS - Small & 1974 (2950) & 3030 (3981) & 5711 (54721) & 478 (582) \\
        SS - Small & 30 (300) & 47 (399) & 39 (5572) & 6 (57) \\
        SS - Medium & 9 (244) & 9 (223) & 24 (5184) & 5 (30) \\
        SS - Large & 6 (191) & 6 (157) & 27 (4422) & 5 (21) \\
        \bottomrule
    \end{tabular}
    \label{tbl:dataset-breakdown}
\end{table}

Table~\ref{tbl:dataset-breakdown} illustrates clear differences between the large-sample and small-sample schemes. In most domains, only a small fraction of the collected SB data tuples ultimately contribute to the training because each tuple expands into many candidate-variable observations. An exception is the FL domain, where each SB data tuple contains relatively few candidates, so most of the collected data is retained and used.
In the small-sample setting, we continue sampling until we reach 25,000 candidate observations. As the problem size increases, each node yields more branching candidates, so fewer subproblems are needed to reach this target. Moreover, unlike the large-sample scheme, we retain all collected observations, which makes better use of each solved instance. As a result, the small-sample setting requires solving orders of magnitude fewer instances to assemble training data.

\subsubsection{Preprocessing}

Before training the sparse regression models, we apply a standard preprocessing pipeline to streamline the feature space. In particular, we remove features that exhibit near-constant variance across the training set. Next, using the remaining features, we construct quadratic expansions by adding squared terms and pairwise interaction products to the list of features.

\section{Computational Results}

We next examine the structure and performance of the resulting sparse quadratic models. We denote by QL and QS the models trained under the large-sample and small-sample settings, respectively.

\subsection{Model sizes}

We report average model size, measured as the number of nonzero parameters. Table~\ref{tbl:sizes} presents the arithmetic mean model size for each problem domain, averaged over five independently generated training datasets. All models in this section are trained using data from small instances. 

\begin{table}[ht]
\caption{Average model sizes for the four problem domains. All models are built using training data from small problems.}
    \centering
    \begin{tabular}{l c c c c}
        \toprule
        Model & SC & CA & FL & IS\\
        \midrule
        QL & 1630 & 1708 & 2043 & 1282\\
        QS & 533 & 633 & 620 & 297\\
        GNN & 64,000 & 64,000 & 64,000 & 64,000\\
        \bottomrule
    \end{tabular}
    \label{tbl:sizes}
\end{table}

Table~\ref{tbl:sizes} contrasts the complexity of the sparse quadratic models with that of the GNN baselines. Across all domains, the sparse models use fewer than 4\% as many parameters as the corresponding GNNs.

With model size established, we next evaluate predictive performance on test data using two complementary test sets, both constructed under the large-sample scheme. 

The \emph{SB test set} contains 20,000 observations, each a complete SB data tuple $(\textbf{G}, \mathbf{i}^*_{\text{SB}}, \textbf{X}, \textbf{s})$, and is used to evaluate performance at the node level. The \emph{candidate test set} contains 100,000 candidate-level observations from small instances, where each observation corresponds to a single candidate variable, such as one row of $\textbf{X}$. Together, these two test capture performance from two angles: node-level selection quality and variable-level score prediction.

\subsection{Model accuracy}
Table~\ref{tbl:acc-results} reports average accuracy for our models and the GNN on the SB test set. We define accuracy as the fraction of SB data tuples for which the model's top-ranked candidate matches SB's top-ranked candidate. Since the SB test set is constructed from small instances, we evaluate models trained on small-instance data.

\begin{table}[ht]
\caption{Accuracy of the models on the SB test set (\%).}
    \centering
    \begin{tabular}{l c c c c}
        \toprule
        Model & SC & CA & FL & IS\\
        \midrule
        QL & 57.8 & 57.6 & 70.1 & 49.0\\
        QS & 57.1 & 56.9 & 68.0 & 46.0\\
        GNN & 62.4 & 58.5 & 72.4 & 54.8\\
        \bottomrule
    \end{tabular}
\label{tbl:acc-results}
\end{table}

Overall, the GNN achieves the highest accuracy, consistent with the greater expressive power of deep networks. However, despite using fewer than 4\% as many parameters, the sparse models remain competitive on the combinatorial auctions (CA) and facility location (FL) domains. 
Finally, models trained with large samples are typically slightly more accurate than those trained in the small-sample setting. Even so, while these results indicate that GNNs better predict SB decisions on small instances, prediction accuracy alone does not determine solver performance, especially as problem size grows.

\subsection{Performance analysis}
We therefore turn to computational experiments that evaluate all learned branching rules, along with SCIP's built-in strategies, on benchmark instances.

\textbf{Baselines.} For context, we compare against several standard branching methods:
\begin{itemize}
    \item  SCIP's default branching rule \textit{hybrid} (\textit{reliability pseudocost) branching} (SCIP). This strategy combines pseudocosts with heuristic safeguards and is the default in state-of-the-art MIP solvers.
    \item  \textit{Vanilla full strong branching} (VFS). A computationally expensive strategy empirically known to produce small search trees. VFS serves as our target for learning.
    \item  GNN-based rules (GNN). This is the leading ML-based branching approach for SB. We present two versions: GNN-C and GNN-G. In both cases, the MIP solver is run on a CPU. In GNN-C, the GNN inference is executed on the CPU. In GNN-G, the GNN inference is executed on a high-end GPU.
\end{itemize}
For the sparse models, we consider two training schemes: (i) large-sample models (QL) trained on small instances and evaluated across all sizes, and (ii) small-sample models (QS) trained and evaluated separately for each problem size.

\textbf{Settings.} We evaluate every method across all problem domains and sizes. For each configuration, we solve 20 benchmark instances under five different random seeds to reduce variance \citep{lodi13}. We treat each instance-seed combination as a separate problem, yielding 100 problems per method. 

Across all runs, we impose a one-hour time limit, disable presolve restarts, and restrict cutting-plane generation to the root node. For VFS, we also set the non-default SCIP parameters \texttt{idempotent = TRUE} and \texttt{scoreall = TRUE} to match the data-collection environment.

Experiments are run on Linux servers equipped with Intel Xeon Gold 6226 CPUs at 2.70 GHz \citep{PACE}. GNN-G additionally utilizes an NVIDIA Tesla V100 (16 GB) GPU for inference. All MIP solves run in a single-threaded mode and in isolation to avoid interference from other jobs. 

We implement sparse models and SCIP's built-in rules through SCIP's C API and deploy the GNN-based rules using Ecole \citep{prouvost+20} and PyTorch \citep{paszke+19}. 

\subsubsection{Comparative results.}

Table~\ref{tbl:results} summarizes the performance of all branching strategies across problem domains and sizes. The ``Solved'' column reports the number of problems that each method solves within the time limit. Problems that are not solved by any method are omitted from the subsequent calculations. 

The ``Time'' column reports the 1-shifted geometric mean CPU runtime (seconds); runs that hit the one-hour time limit are recorded as 3600 seconds. The ``Nodes'' column reports the 1-shifted geometric mean number of branch-and-bound nodes. For runs that time out, we assign the largest node count observed among solved problems in the same domain and size. The only exception is when a problem is solved by GNN-G but not by GNN-C within one hour: in that case, we assign GNN-C the node count of GNN-G, since both methods induce the same search tree.

VFS consistently produces the smallest search trees, but at a substantial computational cost. Runtimes can be up to 65 times higher than those of the learned branching rules, which limits its practical use.

On small instances, the GNN's higher prediction accuracy generally translates into smaller trees than the other learned methods. At the same time, QL attains similar tree sizes while delivering the fastest overall runtimes, outperforming even GNN-G despite GPU-based inference. Models trained in the small-sample setting are typically slightly weaker than their large-sample counterparts, but the differences are modest, suggesting that strong performance is attainable with limited training data.

On medium SC and CA problems, GNN methods still tend to yield the smallest trees. Yet, sparse models often solve faster, demonstrating a key trade-off: faster branching decisions can outweigh modest increases in tree size. For IS, the small-sample models outperform the large-sample ones, consistent with the advantage of training on data that matches the evaluation size regime.

The benefits of small-sample training are most pronounced on large instances. For large CA, GNN-C often reduces tree size, but sparse models remain faster in runtime. For large IS, QS solves more problems than QL and runs significantly faster. Overall, these results support size-matched training as a simple and effective way to improve generalization.

SCIP's default solver, Reliability Pseudocost Branching, occasionally invokes strong branching with side effects enabled. This can lead to understated node counts relative to the true search effort (see \citet{gamrath18} for discussion of fair node counting). For this reason, we avoid direct node-count comparisons between SCIP and the learned methods. 

Runtime comparisons, however, remain informative. Across all problem domains and sizes, the sparse quadratic models consistently outperform SCIP on average, regardless of whether they are trained in the large-sample or small-sample setting. In particular, on large CA instances, QS delivers a 25\% speedup relative to SCIP. In addition to runtime comparisons, maximum memory usage provides an important efficiency metric. Runs using our models consume up to 10\% more memory than SCIP, whereas GNN runs require 5–15 times as much, highlighting the relatively low memory overhead of our approach.

Overall, these results indicate that sparse quadratic branching models can materially improve SCIP's time-to-solution under its standard configuration, delivering speedups while preserving robust solver behavior.

\renewcommand{\arraystretch}{1.1} 
\setlength{\tabcolsep}{3.5pt} 

\begin{table}
\caption{Performance of branching rules on the evaluation instances.}\label{tbl:results}
\small
\begin{subtable}[t]{\textwidth}
    \captionsetup{font=normalsize}
    \centering
    \begin{tabularx}{\textwidth}{
        >{\raggedright\arraybackslash}p{1.2cm}
        *{3}{>{\centering\arraybackslash}X} |
        *{3}{>{\centering\arraybackslash}X} |
        *{3}{>{\centering\arraybackslash}X}
    }
    \toprule
    & \multicolumn{3}{c|}{Small Problems} 
    & \multicolumn{3}{c|}{Medium Problems} 
    & \multicolumn{3}{c}{Large Problems} \\
    \cmidrule(lr){2-10} Method & Solved & Time & Nodes & Solved & Time & Nodes & Solved & Time & Nodes \\
    \midrule
    \midrule
    SCIP & 100 & 6.7 & 55 & 100 & 60 & 2381 & 65 & 1281 & 89785 \\
    VFS & 100 & 37 & 123 & 70 & 1027 & 2552 & 0 & 3600 & 238598 \\
    QL & 100 & 3.7 & 161 & 100 & 40 & 2337 & 69 & 1196 & 74019 \\
    QS & 100 & 3.8 & 170 & 100 & 41 & 2479 & 67 & 1205 & 75031 \\
    GNN-C & 100 & 7.3 & 137 & 100 & 125 & 1932 & 19 & 3250 & 126884 \\
    GNN-G & 100 & 4.4 & 137 & 100 & 48 & 1932 & 63 & 1515 & 59478 \\
    \bottomrule
    \end{tabularx}
    \caption{Set Covering}
\end{subtable}\vfill
\begin{subtable}[t]{\textwidth}
    \captionsetup{font=normalsize}
    \centering
    \begin{tabularx}{\textwidth}{
        >{\raggedright\arraybackslash}p{1.2cm}
        *{3}{>{\centering\arraybackslash}X} |
        *{3}{>{\centering\arraybackslash}X} |
        *{3}{>{\centering\arraybackslash}X}
    }
    \midrule
    \midrule
    SCIP & 100 & 2.5 & 11 & 100 & 18 & 658 & 100 & 137 & 9110 \\
    VFS & 100 & 7.9 & 61 & 100 & 247 & 557 & 24 & 3316 & 44302 \\
    QL & 100 & 1.4 & 74 & 100 & 8.8 & 770 & 100 & 110 & 10139 \\
    QS & 100 & 1.4 & 75 & 100 & 8.6 & 775 & 100 & 103 & 9572 \\
    GNN-C & 100 & 1.7 & 71 & 100 & 13 & 703 & 100 & 160 & 8937 \\
    GNN-G & 100 & 1.9 & 71 & 100 & 11 & 703 & 100 & 120 & 8937 \\
    \bottomrule
    \end{tabularx}
    \caption{Combinatorial Auctions}
\end{subtable}\vfill
\begin{subtable}[t]{\textwidth}
    \captionsetup{font=normalsize}
    \centering
    \begin{tabularx}{\textwidth}{
        >{\raggedright\arraybackslash}p{1.2cm}
        *{3}{>{\centering\arraybackslash}X} |
        *{3}{>{\centering\arraybackslash}X} |
        *{3}{>{\centering\arraybackslash}X}
    }
    \midrule
    \midrule
    SCIP & 100 & 34 & 25 & 100 & 189 & 138 & 95 & 633 & 126 \\
    VFS & 100 & 60 & 86 & 95 & 430 & 257 & 81 & 1524 & 290 \\
    QL & 100 & 28 & 111 & 100 & 144 & 315 & 95 & 589 & 371 \\
    QS & 100 & 29 & 122 & 100 & 147 & 322 & 95 & 561 & 348 \\
    GNN-C & 100 & 33 & 107 & 100 & 182 & 336 & 94 & 717 & 381 \\
    GNN-G & 100 & 29 & 107 & 100 & 158 & 336 & 95 & 640 & 381 \\
    \bottomrule
    \end{tabularx}
    \caption{Capacitated Facility Location}
\end{subtable}\vfill
\begin{subtable}[t]{\textwidth}
    \captionsetup{font=normalsize}
    \centering
    \begin{tabularx}{\textwidth}{
        >{\raggedright\arraybackslash}p{1.2cm}
        *{3}{>{\centering\arraybackslash}X} |
        *{3}{>{\centering\arraybackslash}X} |
        *{3}{>{\centering\arraybackslash}X}
    }
    \midrule
    \midrule
    SCIP & 100 & 6.4 & 19 & 100 & 113 & 2435 & 41 & 1173 & 21163 \\
    VFS & 98 & 111 & 44 & 14 & 2999 & 45523 & 0 & 3600 & 96114 \\
    QL & 100 & 3.8 & 42 & 100 & 67 & 1586 & 34 & 974 & 17274 \\
    QS & 100 & 3.9 & 48 & 100 & 45 & 1148 & 40 & 500 & 10047 \\
    GNN-C & 100 & 4.1 & 44 & 100 & 105 & 2491 & 20 & 1982 & 42451 \\
    GNN-G & 100 & 4.2 & 44 & 100 & 86 & 2491 & 20 & 1818 & 42459 \\
    \bottomrule
    \end{tabularx}
    \caption{Maximum Independent Set}
\end{subtable}
\end{table}

\renewcommand{\arraystretch}{1.0} 
\setlength{\tabcolsep}{6pt} 

\subsubsection{Parametric analysis of sparse quadratic models.}

Sparse quadratic models trained in the large-sample setting typically retain over 1000 nonzero coefficients. We therefore examine how model size affects performance by limiting the number of active coefficients to 25, 50, 100, 500, and 1000 coefficients using the \texttt{dfmax} option in \texttt{glmnet}. This constraint is approximate: the fitted model can exceed the specified limit by a small number of parameters. 

Table~\ref{tbl:parametric} reports the results. Here, QL denotes the unrestricted models, and QL-$x$ denotes models trained with \texttt{dfmax} set to $x$.

\renewcommand{\arraystretch}{1.1}
\setlength{\tabcolsep}{3.5pt}

\begin{table}
\caption{Parametric analysis of the quadratic models.}\label{tbl:parametric}
\small
\begin{subtable}[t]{\textwidth}
    \captionsetup{font=normalsize}
    \centering
    \begin{tabularx}{\textwidth}{
        >{\raggedright\arraybackslash}p{1.2cm}
        *{3}{>{\centering\arraybackslash}X} |
        *{3}{>{\centering\arraybackslash}X} |
        *{3}{>{\centering\arraybackslash}X}
    }
    \toprule
    & \multicolumn{3}{c|}{Small Problems} 
    & \multicolumn{3}{c|}{Medium Problems} 
    & \multicolumn{3}{c}{Large Problems} \\
    \cmidrule(lr){2-10} Method & Solved & Time & Nodes & Solved & Time & Nodes & Solved & Time & Nodes \\
    \midrule
    \midrule
    QL & 100 & 3.7 & 161 & 100 & 40 & 2337 & 69 & 1196 & 74019 \\
    QL-1K & 100 & 3.7 & 161 & 100 & 39 & 2341 & 66 & 1205 & 74199 \\
    QL-500 & 100 & 3.7 & 161 & 100 & 40 & 2378 & 66 & 1229 & 76180 \\
    QL-100 & 100 & 3.8 & 167 & 100 & 43 & 2600 & 57 & 1452 & 90798 \\
    QL-50 & 100 & 3.8 & 171 & 100 & 43 & 2637 & 60 & 1419 & 88709 \\
    QL-25 & 100 & 3.8 & 172 & 100 & 44 & 2715 & 59 & 1464 & 92117 \\
    \bottomrule
    \end{tabularx}
    \caption{Set Covering}
\end{subtable}\vfill
\begin{subtable}[t]{\textwidth}
    \captionsetup{font=normalsize}
    \centering
    \begin{tabularx}{\textwidth}{
        >{\raggedright\arraybackslash}p{1.2cm}
        *{3}{>{\centering\arraybackslash}X} |
        *{3}{>{\centering\arraybackslash}X} |
        *{3}{>{\centering\arraybackslash}X}
    }
    \midrule
    \midrule
    QL & 100 & 1.4 & 74 & 100 & 8.8 & 770 & 100 & 110 & 10139 \\
    QL-1K & 100 & 1.4 & 74 & 100 & 8.8 & 772 & 100 & 108 & 9923 \\
    QL-500 & 100 & 1.4 & 76 & 100 & 8.8 & 773 & 100 & 112 & 10332 \\
    QL-100 & 100 & 1.4 & 77 & 100 & 8.8 & 790 & 100 & 112 & 10315 \\
    QL-50 & 100 & 1.4 & 77 & 100 & 8.9 & 812 & 100 & 116 & 10689 \\
    QL-25 & 100 & 1.4 & 78 & 100 & 9.0 & 830 & 100 & 118 & 11003 \\
    \bottomrule
    \end{tabularx}
    \caption{Combinatorial Auctions}
\end{subtable}\vfill
\begin{subtable}[t]{\textwidth}
    \captionsetup{font=normalsize}
    \centering
    \begin{tabularx}{\textwidth}{
        >{\raggedright\arraybackslash}p{1.2cm}
        *{3}{>{\centering\arraybackslash}X} |
        *{3}{>{\centering\arraybackslash}X} |
        *{3}{>{\centering\arraybackslash}X}
    }
    \midrule
    \midrule
    QL & 100 & 28 & 111 & 100 & 144 & 315 & 95 & 589 & 371 \\
    QL-1K & 100 & 28 & 111 & 100 & 147 & 324 & 95 & 589 & 376 \\
    QL-500 & 100 & 28 & 112 & 100 & 147 & 320 & 95 & 589 & 372 \\
    QL-100 & 100 & 30 & 124 & 100 & 150 & 330 & 95 & 594 & 369 \\
    QL-50 & 100 & 30 & 125 & 100 & 154 & 338 & 95 & 591 & 371 \\
    QL-25 & 100 & 31 & 128 & 100 & 156 & 343 & 95 & 585 & 367 \\
    \bottomrule
    \end{tabularx}
    \caption{Capacitated Facility Location}
\end{subtable}\vfill
\begin{subtable}[t]{\textwidth}
    \captionsetup{font=normalsize}
    \centering
    \begin{tabularx}{\textwidth}{
        >{\raggedright\arraybackslash}p{1.2cm}
        *{3}{>{\centering\arraybackslash}X} |
        *{3}{>{\centering\arraybackslash}X} |
        *{3}{>{\centering\arraybackslash}X}
    }
    \midrule
    \midrule
    QL & 100 & 3.8 & 42 & 100 & 67 & 1586 & 34 & 728 & 11604 \\
    QL-1K & 100 & 3.8 & 43 & 100 & 71 & 1691 & 30 & 849 & 13749 \\
    QL-500 & 100 & 3.8 & 45 & 100 & 81 & 1951 & 27 & 1233 & 20391 \\
    QL-100 & 100 & 3.8 & 45 & 100 & 90 & 2266 & 30 & 1078 & 17324 \\
    QL-50 & 100 & 4.0 & 49 & 100 & 95 & 2427 & 29 & 1061 & 17486 \\
    QL-25 & 100 & 4.0 & 52 & 100 & 87 & 2277 & 32 & 861 & 14573 \\
    \bottomrule
    \end{tabularx}
    \caption{Maximum Independent Set}
\end{subtable}
\end{table}

\renewcommand{\arraystretch}{1.0}
\setlength{\tabcolsep}{6pt}

Table~\ref{tbl:parametric} reports the performance of the size-constrained models on the evaluation instances. Overall, reducing model size tends to degrade performance, both in runtime and in search-tree size, with the effect most pronounced on large instances. For example, on large SC problems, the unrestricted QL models solve ten more instances than QL-25.

\subsubsection{Quadratic versus linear models.}

We next assess whether quadratic feature expansions are important for performance. To isolate their effect, we train large-sample sparse models without quadratic terms (LL). Table~\ref{tbl:quadvslinear} compares these linear models to the corresponding quadratic QL models using the same evaluation metrics.

\begin{table}[!htbp]
\caption{Performance of branching rules with quadratic and linear features.}\label{tbl:quadvslinear}
\small
\begin{subtable}[t]{\textwidth}
    \captionsetup{font=normalsize}
    \centering
    \begin{tabularx}{\textwidth}{
        >{\raggedright\arraybackslash}p{1.2cm}
        *{3}{>{\centering\arraybackslash}X} |
        *{3}{>{\centering\arraybackslash}X} |
        *{3}{>{\centering\arraybackslash}X}
    }
    \toprule
    & \multicolumn{3}{c|}{Small Problems} 
    & \multicolumn{3}{c|}{Medium Problems} 
    & \multicolumn{3}{c}{Large Problems} \\
    \cmidrule(lr){2-10} Method & Solved & Time & Nodes & Solved & Time & Nodes & Solved & Time & Nodes \\
    \midrule
    \midrule
    QL & 100 & 3.7 & 161 & 100 & 40 & 2337 & 69 & 1196 & 74019 \\
    LL & 100 & 3.8 & 181 & 100 & 49 & 3087 & 45 & 1915 & 125469 \\
    \bottomrule
    \end{tabularx}
    \caption{Set Covering}
\end{subtable}\vfill
\begin{subtable}[t]{\textwidth}
    \captionsetup{font=normalsize}
    \centering
    \begin{tabularx}{\textwidth}{
        >{\raggedright\arraybackslash}p{1.2cm}
        *{3}{>{\centering\arraybackslash}X} |
        *{3}{>{\centering\arraybackslash}X} |
        *{3}{>{\centering\arraybackslash}X}
    }
    \midrule
    \midrule
    QL & 100 & 1.4 & 74 & 100 & 8.8 & 770 & 100 & 110 & 10139 \\
    LL & 100 & 1.4 & 79 & 100 & 9.3 & 875 & 100 & 153 & 14566 \\
    \bottomrule
    \end{tabularx}
    \caption{Combinatorial Auctions}
\end{subtable}\vfill
\begin{subtable}[t]{\textwidth}
    \captionsetup{font=normalsize}
    \centering
    \begin{tabularx}{\textwidth}{
        >{\raggedright\arraybackslash}p{1.2cm}
        *{3}{>{\centering\arraybackslash}X} |
        *{3}{>{\centering\arraybackslash}X} |
        *{3}{>{\centering\arraybackslash}X}
    }
    \midrule
    \midrule
    QL & 100 & 28 & 111 & 100 & 144 & 315 & 95 & 589 & 371 \\
    LL & 100 & 30 & 128 & 100 & 152 & 337 & 95 & 587 & 364 \\
    \bottomrule
    \end{tabularx}
    \caption{Capacitated Facility Location}
\end{subtable}\vfill
\begin{subtable}[t]{\textwidth}
    \captionsetup{font=normalsize}
    \centering
    \begin{tabularx}{\textwidth}{
        >{\raggedright\arraybackslash}p{1.2cm}
        *{3}{>{\centering\arraybackslash}X} |
        *{3}{>{\centering\arraybackslash}X} |
        *{3}{>{\centering\arraybackslash}X}
    }
    \midrule
    \midrule
    QL & 100 & 3.8 & 42 & 100 & 67 & 1586 & 34 & 663 & 10426 \\
    LL & 100 & 4.8 & 63 & 94 & 270 & 7653 & 7 & 3007 & 52182 \\
    \bottomrule
    \end{tabularx}
    \caption{Maximum Independent Set}
\end{subtable}
\end{table}

Table~\ref{tbl:quadvslinear} indicates that quadratic feature transformations are important for performance on medium and large SC, CA, and IS problems. On small instances, linear and quadratic models perform similarly, although quadratic models typically yield smaller search trees. As instances grow in size and complexity, the gap widens: for medium and large-scale SC and IS problems, quadratic models are not only faster but also solve more instances than their linear counterparts. Overall, the SC and IS results suggest that linear models alone are insufficient to learn effective branching-score predictors.

\subsubsection{Interpretation of quadratic models.}

Because the learned quadratic models contain relatively few active terms, they are amenable to interpretation and systematic analysis. We illustrate this using the QL-500 models, the smallest variant that remains predictive, by examining which features are selected most often and which terms and coefficients appear most influential. 

Recall that we solve five models for each of the four problem classes. The following four features were selected in all 20 models:

\begin{itemize}
\item minimum row RHS squared
\item integral violation $\times$ maximum weighted active coefficient (type 4)
\item ceiling distance $\times$ solution value
\item ceiling distance $\times$ floor distance
\end{itemize}
The fact that these features appear across all models suggests that they capture drivers of strong branching that persist across problem types.

For meaningful comparisons of importance between features of different magnitudes, we scale all model coefficients using standardized coefficients, i.e., we set $\beta_i^s = \hat{\beta}_i s_i / s_y$, where $s_i$ and $s_y$ are the standard deviations of feature $i$ and outcome $y$ in the training dataset, respectively.

\begin{figure}[!htbp]
    \caption{Heat map of features with largest absolute coefficients in QL-500 models.}
    \includegraphics[width=\textwidth]{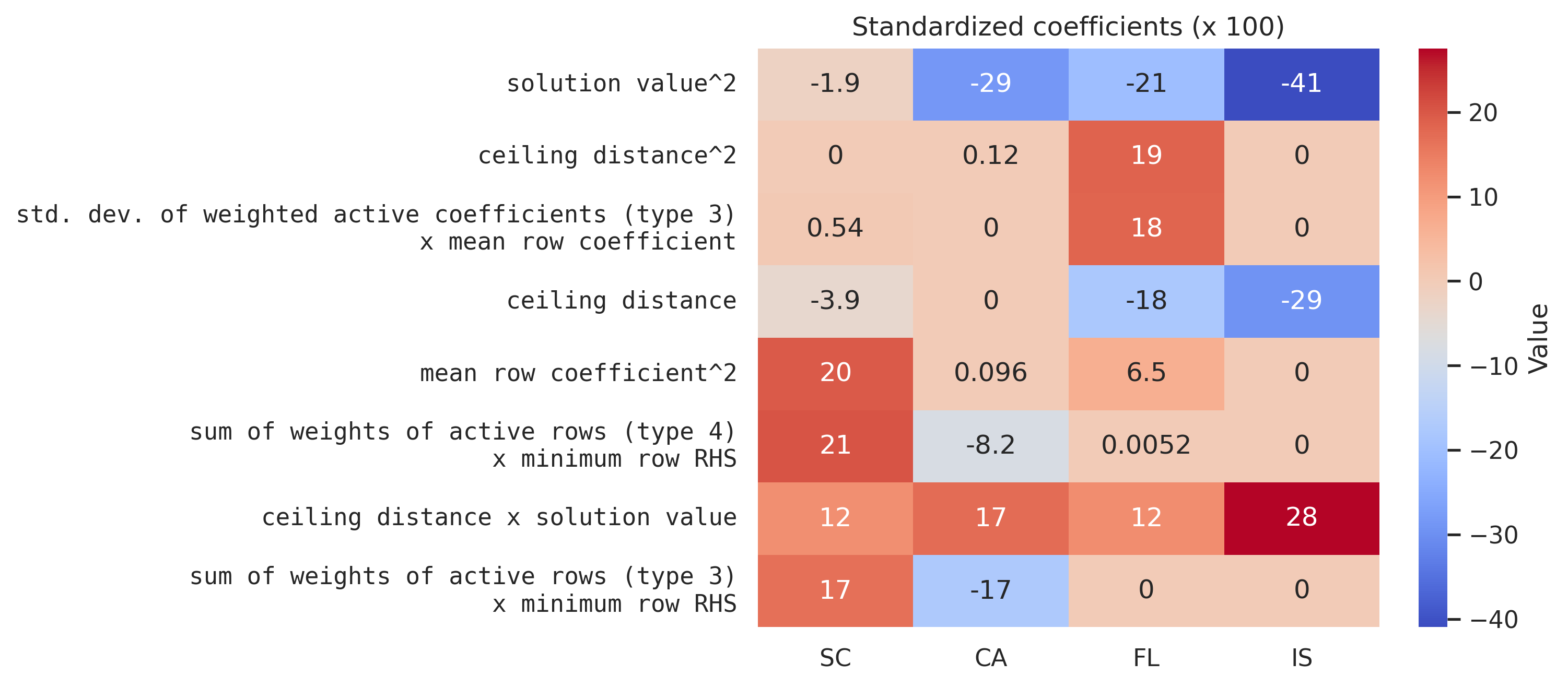}
    \label{fig:heatmap_l}
\end{figure}

Figure \ref{fig:heatmap_l} reports, for each of the four domains, the three features with the largest coefficients in absolute value. Several terms, for example, solution value$^2$, receive large coefficients in multiple domains, indicating broad relevance. Others are more domain-specific and reflect structural differences between problem classes. For instance, the mean row coefficient$^2$ plays a prominent role for SC but is not similarly important for IS. 

\section{Conclusions}

Across four well-studied problem classes, we demonstrate that sparse quadratic models for predicting strong branching scores can yield faster branch-and-bound performance than both SCIP's default strategy and state-of-the-art GNN-based branching models. Although GNNs achieve higher strong branching score predictive accuracy and produce smaller search trees, their training and deployment come with high computational costs. Whereas GNNs are not competitive without a GPU, sparse quadratic models are highly efficient. They require only a fraction of the computational resources and training data needed by deep learning models, yet still deliver strong performance. This combination of data efficiency and low deployment cost makes sparse quadratic models an attractive choice for real-world solver integration, particularly in settings where GPU access and training budgets are limited.

\setlength{\tabcolsep}{3pt}
\renewcommand{\arraystretch}{1.1}

\begin{center}
\LARGE{\textbf{Appendix}}
\end{center}

\appendix
\section{Features}

Table~\ref{tbl:features} lists all features used in this study; features introduced by \citet{khalil+16} and \citet{gasse+19} are labeled ``K'' and ``G'', respectively. After exploratory experimentation, we decided to combine these two sets of features; using either one of them alone results in somewhat inferior results. We use min., max., and std. dev. abbreviations for minimum, maximum, and standard deviation, respectively. As in \citet{khalil+16}, we apply min-max normalization to scale the K features in each node to the [0,1] range.

\begin{longtable}{p{3.5cm} c c p{9.5cm}}
\caption{Features used for building sparse models}\\ 
\toprule
\textbf{Feature} & \textbf{Source} & \textbf{Count} & \textbf{Description} \\
\midrule
\endfirsthead
\caption{continued} \\
\textbf{Feature} & \textbf{Source} & \textbf{Count} & \textbf{Description} \\
\midrule
\endhead
Objective coefficient & K & 3 & Value of the coefficient (raw, positive only, negative only)\\
Number of constraints & K & 1 & The number of constraints in which the candidate appears\\
Statistics for constraint degrees & K & 4 & The mean, std. dev., min., and max. of the constraint degrees (the number of variables with nonzero coefficients)\\
Statistics for positive (negative) constraint coefficients & K & 10 & The count, mean, std. dev., min., and max. of the candidate's positive (negative) coefficients in the constraints, normalized\\
Integrality violation & K & 1 & $\min(\hat{x}_j - \floor{\hat{x}_j}, \ceil{\hat{x}_j} - \hat{x}_j)$\\
Ceiling distance & K & 1 & $\ceil{\hat{x}_j} - \hat{x}_j$\\
Floor distance & G & 1 & $\hat{x}_j - \floor{\hat{x}_j}$\\
Weighted pseudocosts & K & 5 & Up and down pseudocosts weighted by the ceiling and floor distances of $\hat{x}_j$, respectively, and their ratio, sum, and product\\
Infeasibility statistics & K & 4 & Number and fraction of times that branching on a variable, in either direction, led to immediate pruning of the resulting child node\\
\raggedright Min. and max. of ratios of constraint coefficients to right-hand-sides (RHS) & K & 4 & Min. and max. of the ratios of the constraint coefficients to positive (negative) RHSs\\
Min. and max. of one-to-all coefficient ratios & K & 8 & The ratio of the candidate's (positive/negative) coefficient to the sum of coefficients with the same or opposite sign in the same constraint (4 cases: pos/pos, pos/neg, neg/pos, neg/neg)\\
Statistics for active constraint coefficients & K & 24 & Active constraints at the LP solution (i.e., those satisfied at equality) are assigned four types of weights: ($i$) unit weight; ($ii$) inverse of the sum of the absolute coefficients of all variables in the constraint, ($iii$) inverse of the sum of the  absolute coefficients of all candidates in the constraint, and ($iv$) absolute value of the constraint's dual value. For each weighting scheme, we compute the sum, mean, std. dev., max., and min. of the weighted absolute coefficients, as well as the weighted count of active constraints.\\
Type & G & 4 & Type (binary, integer, implied integer, continuous)\\
Has lower (upper) bound & G & 2 & Whether the candidate has a lower (upper) bound\\
Solution is at lower (upper) bound & G & 2 & Whether the candidate's LP solution value equals its lower (upper) bound\\
Basis status & G & 4 & Simplex basis status (lower, basic, upper, zero)\\
Reduced cost & G & 1 & Reduced cost, normalized\\
LP age & G & 1 & Number of successive times the candidate was in the LP and attained a value of zero in the LP solutions\\
Solution value & G & 1 & Solution value of the candidate at the LP solution\\
Incumbent value & G & 1 & Solution value of candidate in the incumbent solution\\
Average incumbent value & G & 1 & Average value of candidate in the incumbent solutions\\
Statistics for constraint coefficients & G & 3 & The mean, min., and max.  of the candidate's normalized coefficients across the constraints in which it appears\\
Statistics for objective cosine similarity & G & 3 & The mean, min., and max. of the cosine similarities between the objective function and each constraint involving the candidate\\
Statistics for RHSs & G & 3 & The mean, min., and max., of the RHSs of the candidate's constraints, normalized\\
Statistics for constraint tightness & G & 3 & The mean, min. and max. of the tightness indicators (whether the constraint is active) of the candidate's constraints\\
Statistics for dual solutions of constraints & G & 3 & The mean, min., and max., of the dual solution values of the candidate's constraints, normalized\\
Statistics for LP ages of constraints & G & 3 & The mean, min. and max. of the LP ages of the candidate's constraints (number of successive times the constraint was in the LP and was not tight in the solution)\\
\bottomrule
\label{tbl:features}
\end{longtable}

\subsection*{Acknowledgments}
This work was conducted as part of the Institute for the Design of Advanced Energy Systems (IDAES) with support through the Simulation-Based Engineering, Crosscutting Research Program and the Solid Oxide Fuel Cell Program’s Integrated Energy Systems thrust within the U.S. Department of Energy’s Hydrocarbons and Geothermal Energy Office.

This research was supported in part through research cyberinfrastructure resources and services provided by the Partnership for an Advanced Computing Environment (PACE) at the Georgia Institute of Technology, Atlanta, Georgia, US \citep{PACE}.

\subsection*{Disclaimer}
This project was funded by the Department of Energy, National Energy Technology Laboratory an agency of the United States Government, in part, through a site support contract. Neither the United States Government nor any agency thereof, nor any of its employees, nor the support contractor, nor any of their employees, makes any warranty, express or implied, or assumes any legal liability or responsibility for the accuracy, completeness, or usefulness of any information, apparatus, product, or process disclosed, or represents that its use would not infringe privately owned rights. Reference herein to any specific commercial product, process, or service by trade name, trademark, manufacturer, or otherwise does not necessarily constitute or imply its endorsement, recommendation, or favoring by the United States Government or any agency thereof. The views and opinions of authors expressed herein do not necessarily state or reflect those of the United States Government or any agency thereof.

\bibliographystyle{abbrvnat}
\bibliography{main}

\end{document}